\documentclass[9pt,twocolumn,twoside]{pnas-new} 

\usepackage{comment}
\usepackage{url}
\usepackage{hyperref}

\renewcommand{\arraystretch}{1.6} 

\templatetype{pnasresearcharticle} 
\begin{document}
\title{Integrating topic modeling and word embedding to characterize violent deaths}

\author[a,1]{Alina Arseniev-Koehler} 
\author[b]{Susan D. Cochran} 
\author[c]{Vickie M. Mays}
\author[d]{Kai-Wei Chang}
\author[a,1]{Jacob Gates Foster}

\affil[a]{Department of Sociology, University of California-Los Angeles, Los Angeles, CA 90095}
\affil[b]{Department of Epidemiology, UCLA Fielding School of Public Health and Department of Statistics, University of California-Los Angeles, Los Angeles, CA 90095}
\affil[c]{Department of Psychology, Department of Health Policy and Management, UCLA Fielding School of Public Health, and BRITE Center, University of California-Los Angeles, Los Angeles, CA 90095}
\affil[d]{Department of Computer Science, University of California-Los Angeles, Los Angeles, CA 90095}

\leadauthor{Arseniev-Koehler}


\correspondingauthor{\textsuperscript{1}To whom correspondence may be addressed. Email: arsena@ucla.edu; foster@soc.ucla.edu}


\dates{This manuscript was compiled on \today}

\maketitle

\begin{abstract} 
There is an escalating need for methods to identify latent patterns in text data from many domains. We introduce a new method to identify topics in a corpus and represent documents as topic sequences. Discourse Atom Topic Modeling draws on advances in theoretical machine learning to integrate topic modeling and word embedding, capitalizing on the distinct capabilities of each. We first identify a set of vectors (``discourse atoms'') that provide a sparse representation of an embedding space. Atom vectors can be interpreted as latent topics: Through a generative model, atoms map onto distributions over words; one can also infer the topic that generated a sequence of words. We illustrate our method with a prominent example of underutilized text: the U.S. National Violent Death Reporting System (NVDRS). The NVDRS summarizes violent death incidents with structured variables and unstructured narratives. We identify 225 latent topics in the narratives (e.g., preparation for death and physical aggression); many of these topics are not captured by existing structured variables. Motivated by known patterns in suicide and homicide  by gender, and recent research on gender biases in semantic space, we identify the gender bias of our topics (e.g., a topic about pain medication is feminine). We then compare the gender bias of topics to their prevalence in narratives of female versus male victims. Results provide a detailed quantitative picture of reporting about lethal violence and its gendered nature. Our method offers a flexible and broadly applicable approach to model topics in text data.

\end{abstract}

\ifthenelse{\boolean{shortarticle}}{\ifthenelse{\boolean{singlecolumn}}{\abscontentformatted}{\abscontent}}{}

\dropcap{D}igital technologies have produced a deluge of computer-readable text: tweets, blogs, scientific articles, product reviews, legal documents, financial reports, electronic health records, and administrative documents (e.g., those used in public health surveillance). Despite the promise of such rich data, deriving meaningful information from large-scale text remains a challenge \cite{hirschberg2015advances}. This is especially so in real-world applications, which often put particular demands on computational text analysis methods. Such methods should be interpretable. They should adapt to the nuances of different corpora and discourses. And they should have strong theoretical foundations. In this paper, we offer a new approach that meets these demands: Discourse Atom Topic Modeling (DATM). DATM integrates topic modeling \citep{blei2012probabilistic} and word embedding \citep{mikolov2013efficient} to identify latent topics in embeddings and map documents onto topics. Methods developed for embeddings \cite[e.g., latent dimensions of cultural meaning in][]{garg2018word, bolukbasi2016quantifying} can be applied directly to the topics. We illustrate the value of DATM using text data collected as part of a large, ongoing, surveillance system for lethal violence in the U.S.

Violent death surveillance provides a striking example of the promises and challenges of computational text analysis. Violent deaths (e.g., suicide and homicide) are among the leading causes of mortality in the U.S. \cite{murphy2018mortality}: more than seven people per hour die a violent death \cite{nvdrshomepage}.  Understanding and reducing the frequency of these deaths is a major goal for public health. Much of what we know about violent death comes from large administrative databases like the NVDRS, a nationwide public-health surveillance database established by the Centers for Disease Control in 2003 \citep[e.g.,][]{ertl2019surveillance, paulozzi2004cdc}. The NVDRS contains both structured variables (e.g., victim demographics) and unstructured text narratives. These narratives describe the circumstances of death incidents based on reports from law enforcement, medical examiners/coroners, toxicology reports, and crime laboratories. While much has been learned from the NVDRS since its inception, researchers have largely used the structured measures; traditional qualitative methods are too labor-intensive to use at scale. The narratives, summarizing more than 300,000 violent deaths, remain mostly unused, despite their potential to illuminate many aspects of violent death, from proximate correlates to nuanced context.

One well-known and durable pattern that may manifest in text descriptions is differences in violence by gender. Men are more likely than women to die from and perpetrate lethal violence \citep{batton2004gender, fox2017gender}. Men and women victims also tend to die by different methods \citep{callanan2012gender, fox2017gender}. Among suicides and homicides, for example, men are more likely to use firearms, while women are more likely to use poisonous substances \citep{fox2017gender, callanan2012gender}. While such gender-linked patterns are reflected across structured variables in the NVDRS (and are well-documented in the research literature), the NVDRS narratives may also encode gendered patterns---some as yet unidentified. Indeed, a growing body of computational work illustrates the extent to which information about gender manifests in language \citep[e.g.,][]{bolukbasi2016quantifying, garg2018word}.  

The case of violent death surveillance highlights two problems that computational text analysis can solve. First, researchers often want to summarize large corpora, e.g. by extracting major themes like  ``hot'' scientific topics in PNAS \citep{griffiths2004finding}. Second, researchers want to find evidence for patterns suggested by theory or previous scholarship, e.g., the presence and dynamics of gender and ethnic stereotypes in media language \citep{garg2018word}.
Existing methods can solve both of these problems, but separately. DATM enables us to do both at once. It integrates two major innovations in computational text analysis: topic modeling \citep{blei2012probabilistic} and word embedding \citep[e.g.,][]{mikolov2013efficient}.

Topic modeling identifies latent topics in a corpus. In conventional topic modeling, topics are distributions over words, and documents are distributions over topics. Topic modeling methods identify latent themes in a corpus and connect those themes to the manifest words and documents. Powerful as they are, existing topic modeling approaches remain largely disconnected from contemporary strategies to represent semantic information using word embedding \citep[but see][]{dieng2019topic, das2015gaussian}. 

Word embedding methods represent word meanings by mapping each word in the vocabulary to a point in an $\mathit{N}$-dimensional semantic space (a ``word vector''). Words used in similar contexts in the corpus are mapped to points that are close together. In a well-trained embedding, word vectors represent semantic information in ways that correspond to human meanings. For example, words that humans rate as semantically similar tend to be closer in semantic space. While word embeddings explicitly model words, they may also encode latent semantic structures, e.g., latent dimensions that correspond to gendered meanings \citep[e.g.,][]{bolukbasi2016quantifying, arseniev2020machine, garg2018word}; analysts can then use these dimensions to quantify the gendered meanings of all the words in a corpus. Topic modeling and word embedding thus have distinct strengths and limitations.

DATM identifies topics (latent themes) and infers the distribution of topics in a specific document, just like a standard topic model. Unlike standard topic modeling, however, DATM does so in an explicit embedding framework; both words and topics live in one semantic space. Our method therefore offers rich representations of topics, words, phrases, and latent semantic dimensions in language. It is able to do so by combining several theoretical advances to explain word embeddings and related methodological advances to efficiently represent sentences in semantic space \citep{arora2016latent, arora2018linear, arora2016simple, ethayarajh2018unsupervised}. 

After describing DATM, we use it to identify key topics in narratives describing over 300,000 violent deaths in the NVDRS (2003-2017). We observe a range of topics, including ones about family, preparation for death, and about causality. Using recent approaches to identify semantic dimensions in embedding space \citep{bolukbasi2016quantifying,arseniev2020machine} we identify a gender dimension and compute the gendered meanings of our \textit{topics}. We describe two illustrative topics in depth: (1) long guns (e.g., rifles and shotguns) and (2) sedative and pain medications. Our approach therefore enables us to summarize and contextualize large-scale, unstructured accounts of violent death, while also focusing in on ``needles'' in this haystack of data \citep{boyd2014care}.

\section*{Integrating topic modeling and word embedding}

To integrate topic modeling and word embedding, we address two core methodological challenges. First, we identify latent topics in a trained word embedding space (also referred to as semantic space). In our case, we set out to identify latent topics in an embedding space trained on narratives of violent death. Second, we identify the topic(s) underlying an observed set of words (e.g., a sentence, document, or written death narrative). More generally, we need a theoretical framework to connect a trained embedding space to raw text data. DATM integrates several methodological and theoretical advances in research on word embeddings to address these two challenges, as we describe next. Details are in the Materials and Methods and Supplementary Information (SI). 

\subsection*{Identifying Topics in Semantic Space}

We begin with a word embedding trained on a specific corpus (in our case, narratives of violent death). To identify topics in this embedding space, we apply K-SVD, a sparse dictionary learning algorithm \citep{aharon2006k, rubinstein2008efficient}, to the word-vectors \citep{arora2018linear}. This algorithm outputs a set of $K$ vectors \cite[called ``discourse atoms'' by ][]{arora2018linear} such that any of the $V$ word vectors in the vocabulary can be written as a sparse linear combination of atom vectors. Using the generative model below (Equation \ref{eqref:eqn1}), atom vectors can be interpreted as topics in the embedding space. The words closest to each atom vector typify the topic.

We apply K-SVD to our word embedding while varying the number of atom vectors $K$. To select a final sparse representation, we use two previously proposed metrics for topic model quality---coherence \citep{aletras2013evaluating} and diversity \citep{dieng2019topic}---as well as a third metric suitable for K-SVD ($R^2$), which we refer to as coverage. Together, these metrics quantify 1) how internally coherent topics are (coherence); 2) how distinctive topics are from each other (diversity); and 3) how well the underlying atoms explain the semantic space itself (coverage). We select our final model (with 225 topics) to balance performance across these metrics. See SI for a comparison with topics identified using Latent Dirichlet Allocation topic modeling on our data.

\subsection*{Moving from Semantic Space to Text Data, and Back}

Sparse dictionary learning offers a way to identify the ``building blocks'' of semantic space, but it does not map observed sequences of words (e.g., sentences) to these building blocks. Fortunately, a recently proposed language model offers a link between observed words and points in semantic space: the Latent Variable Model \citep{arora2016latent, arora2016simple}. This model provides a theoretically motivated algorithm to summarize a given set of words as a \textit{context vector} in the semantic space, i.e., a sentence embedding \cite{arora2016simple, ethayarajh2018unsupervised}. For a given context vector, we can then find the closest atom vector in semantic space, and thus map observed sequences of text data to latent topics. For each document, we assign each context window in a sequence of context windows to a topic. This yields a sequence (or, ignoring order, a distribution) of latent topics that represent the document. 

\subsubsection*{The Latent Variable Model}

Consider first a simplified version of the Latent Variable Model (Equation \ref{eqref:eqn1}). In this generative model, the probability of a word $w$ being present at some location $t$ in the corpus is based on the similarity between its word vector $\mathbf{w}$ and the latent ``gist'' at that point in the corpus $\mathbf{c}_t$, i.e., the discourse vector \citep{arora2016latent}. The word most likely to appear at $t$ is the word most similar (closest in semantic space) to the current gist.\footnote{Similarity is measured as the dot product between the two vectors.} The similarities between possible word vectors and the discourse vector can be turned into a probability distribution over words by (1) exponentiating the similarities and (2) dividing by their sum $Z_{c_t}$, so that the distribution sums to 1 (Eq. \ref{eqref:eqn1}). The gist is latent; $\mathbf{c}_t$ is a vector in the semantic space that represents the underlying meaning of the current context. Eq. \ref{eqref:eqn1} thus associates a distribution over words to every point in semantic space, and sets up a correspondence between atom vectors and topics. In the generative model \citep{arora2016latent}, the gist makes a slow random walk through semantic space; at each step $t$ a word is emitted according to:   

\begin{equation}\label{eqref:eqn1}
\textrm{Pr}[w \textrm{ emitted at } t | c_t ] = \frac{ \exp(  \langle \mathbf{c}_t, \mathbf{w} \rangle)}{Z_{{c_t}}}. 
\end{equation}
\noindent This simple model is enough to recover many properties of word embeddings \citep{arora2016latent}. 

Arora \textit{et al.} \citep{arora2016simple} build on Equation \ref{eqref:eqn1} to give a more realistic generative model. The conditional probability of a word \textit{w} being present at some point \textit{t} in the corpus depends on several factors. It depends on the overall frequency of the word in the corpus, $p(\textit{w})$. But it also depends on the local context  or ``gist'' (the familiar $\mathbf{c}_t$), as well as the global context of the corpus ($\mathbf{c}_0$). The global context vector $\mathbf{c}_0$ represents the overall syntactic and semantic structure of the corpus, independent of any local context. The specific combination of local and global context vectors $\mathbf{\tilde{c}}_t$ is a linear combination of $\mathbf{c}_t$ and $\mathbf{c}_0$. The relative importance of word frequency and context is controlled by the hyperparameter $\alpha$; local and global context tradeoff with hyperparameter $\beta$. This improved Latent Variable Model is written formally in Equation \ref{eqref:eqn2} and detailed elsewhere \citep{arora2016latent, arora2016simple, arora2018linear, ethayarajh2018unsupervised}.

\begin{equation}\label{eqref:eqn2}
\textrm{Pr}[w\textrm{ emitted at } t | c_t ] = \alpha p(w) + (1-\alpha) \frac{ \exp(  \langle \mathbf{\tilde{c}}_t, \mathbf{w} \rangle)}{Z_{\tilde{c_t}}}   
\end{equation}

\noindent where $\mathbf{\tilde{c}}_t = \beta \mathbf{c}_0 + (1- \beta)\mathbf{c}_t$ and  $\mathbf{c}_0 \perp \mathbf{c}_t$.

\subsubsection*{Mapping Observed Words into Semantic Space}

In the generative direction, Equation \ref{eqref:eqn2} fixes the probability of a word appearing, given details of the corpus and the current gist. In applications, however, we observe the words; the gist is latent. We want to infer the gist (i.e., where we are semantic space) given an observed set of context words. This allows us to attribute context words to an atom vector. We summarize work by Arora and colleagues that use this model to derive a theoretically motivated, high quality embedding of a set of context words: Smooth Inverse Frequency (SIF) embeddings \cite{arora2016latent, arora2016simple}.

Using Equation \ref{eqref:eqn2}, we can find the Maximum a Posteriori estimate of the combined context vector $\mathbf{\tilde{c}}_t$ for a set of context words $C$ \citep[see][]{arora2016latent, arora2016simple}:

\begin{equation}\label{eqref:eqn3} (\mathbf{\tilde{c}}_t)_{\textrm{MAP}} = \sum_{w \in \mathcal{C}  }^{}   \frac{a}{p(w)+a}\mathbf{w} \textrm{, where } a = \frac{1-\alpha}{\alpha Z} . \end{equation}
\noindent $(\mathbf{\tilde{c}}_t)_{\textrm{MAP}}$ is a weighted average of the word vectors in the context window; words are weighted based on their corpus frequency $p(w)$. Frequent words make a smaller contribution to the estimate of $\mathbf{\tilde{c}}_t$.\footnote{The amount of re-weighting is controlled by the parameter $a$. A lower value for $a$ leads to more extensive down-weighting of frequent words, compared to less frequent words. This parameter ranges reasonably from 0.001 to 0.0001; we use a value of 0.001 \cite{arora2016simple, ethayarajh2018unsupervised}.}

For a given set of context words, we now have an estimate of $\mathbf{\tilde{c}}_t$ (recall that $\mathbf{\tilde{c}}_t$ is a linear combination of local gist and global context for the corpus). But we are fundamentally interested in the local gist $\mathbf{c}_t$. To recover this, we need an estimate of the \textit{global context} $\textbf{c}_0$, which we can then subtract from our estimate of $\tilde{\mathbf{c}}_t$; \cite[see][]{arora2016simple}. We first estimate $\tilde{\mathbf{c}}_t$ for a sample of context windows (e.g., sentences) in the data using Equation \ref{eqref:eqn3}. Then we compute the first principal component of the $\mathbf{\tilde{c}}_t$'s, recovering the direction with the most variance among the context vectors. We interpret this first principal component as the global context vector $\mathbf{c_0}$.\footnote{After we subtract off the projection of a specific $\mathbf{\tilde{c}}_t$ onto the global context vector $\mathbf{c_0}$, the remaining vector $\mathbf{c}_t$ captures the local context---the gist of what is being talked about. We have removed any information in $\mathbf{\tilde{c}}_t$ that corresponds to what the corpus as a whole usually talks about, or how it talks about this (i.e., $\mathbf{c}_t$ does not just capture frequent words).} For a given set of context words $C$, we can estimate $\mathbf{c}_t$ by using Equation \ref{eqref:eqn3} to compute $\mathbf{\tilde{c}}_t$ from the word-vectors in $C$, and then subtracting its projection onto $\mathbf{c}_0$. The result is an estimate of the latent gist of $C$, as a point in semantic space.

Prior work \cite{arora2016simple, ethayarajh2018unsupervised} demonstrates that SIF sentence embedding (i.e., weighting word vectors by frequency, summing them, and removing the global context vector) is also an \textit{empirically} effective representation of the meaning captured by a sentence (or other set of words). In fact, by several metrics, SIF embedding outperforms more sophisticated approaches to represent sentences. We also observe that SIF embedding appears to limit the influence of stopwords on the estimate, without our needing to specify a list of stopwords. Readers familiar with word embeddings may note the correspondence between this MAP and representations of context in the Continuous-Bag-of-Words (CBOW) model \citep{mikolov2013efficient, arora2016simple}; see SI.

SIF embedding allows us to map from a set of observed words to a location in semantic space.  Given that location, we can find the atom vector that is most similar to this estimated gist, i.e., $\arg\max_{\mathbf{k} \in \mathcal{K}} \cos \left(  \mathbf{k},  \mathbf{c}_t \right)$. This atom vector $\mathbf{k}'$ then immediately yields the closest \textit{topic} in semantic space. 

We have combined three ingredients---sparse coding of the embedding space \citep{arora2018linear}, the Latent Variable Model \citep{arora2016latent}, and SIF embeddings \citep{arora2016simple}---into a procedure that allows researchers to discover topics (i.e., atom vectors) and to identify the topic that best matches the estimated gist of an observed context window. To infer topics across an entire document, we estimate the gist ($\mathbf{c}_t$) over rolling context windows in the document.\footnote{Here, we use a context window size with 10 terms.} This is consistent with a key assumption of the Latent Variable Model: That the gist changes slowly across a document. This last step yields the sequence (or, ignoring order and dividing the topic counts by a normalizing constant, the \textit{distribution}) of topics underlying the document.\footnote{Turning the sequence into a probability distribution accounts for differences in word count; it also provides a connection to traditional topic modeling, which associates a distribution over topics to each document.} Here, we code topics as binary variables for each record (present or not). 

\section*{Topics in Descriptions of Violent Death}

When we applied DATM to the NVDRS narratives, the resulting 225 topics covered a range of aspects of violent death. For example, we observed several topics about weapons, substance use, forensic analyses, and social institutions (e.g., jails, family). To interpret a given topic, we examine the 25 most representative terms (ordered by cosine similarity) and then we assign the topic a label using face validity. We list several topics in table \ref{table:sampletopics} and provide a full list in the SI.

\begin{table*}[tbhp]
\caption{Sample Topics within narratives of Violent Death}\label{table:sampletopics}
\begin{tabular}{ll}
\textbf{Topic Label} & \textbf{Seven Most Representative Terms}   \\
\midrule
Physical Aggression & tackled, lunged\_toward, began\_attacking, advanced\_toward, attacked, slapped, intervened  \\
    Causal Language & sparked, preceded, triggered, precipitated, led, prompted, culminated \\
  Preparation for Death & disposal, deeds, prepaid\_funeral, burial, worldly, miscellaneous, pawning\\
  Cleanliness & unkempt, messy, disorganized, cluttered, dirty, tidy, filthy\\
  Everything Seemed Fine & fell\_asleep, everything\_seemed\_fine, seemed\_fine, wakes\_up, ran\_errands, ate\_breakfast, watched\_television \\
  Suspicion and Paranoia & conspiring\_against, plotting\_against, restraining\_order\_filed\_against, belittled, please\_forgive, making\_fun, reminded\\
  Reclusive Behavior and Chronic Illness & recluse, heavy\_drinker, very\_ill, chronic\_alcoholic, bedridden, reclusive, recovering\_alcoholic   \\

\midrule
\multicolumn{2}{p{18cm}}{\textit{Notes:} Most representative terms are listed in order of highest to lowest cosine similarity to each topic's atom vector. Topic labels are manually assigned.} \\

\end{tabular}
\end{table*}

\begin{figure}[tbhp]
\caption{Prevalence of 225 Topics in Narratives of 272,964 Decedents of Violent Death, by Manner of Death.}\label{figure:Heatmap of Topics by MOD}
\includegraphics[scale=0.6]{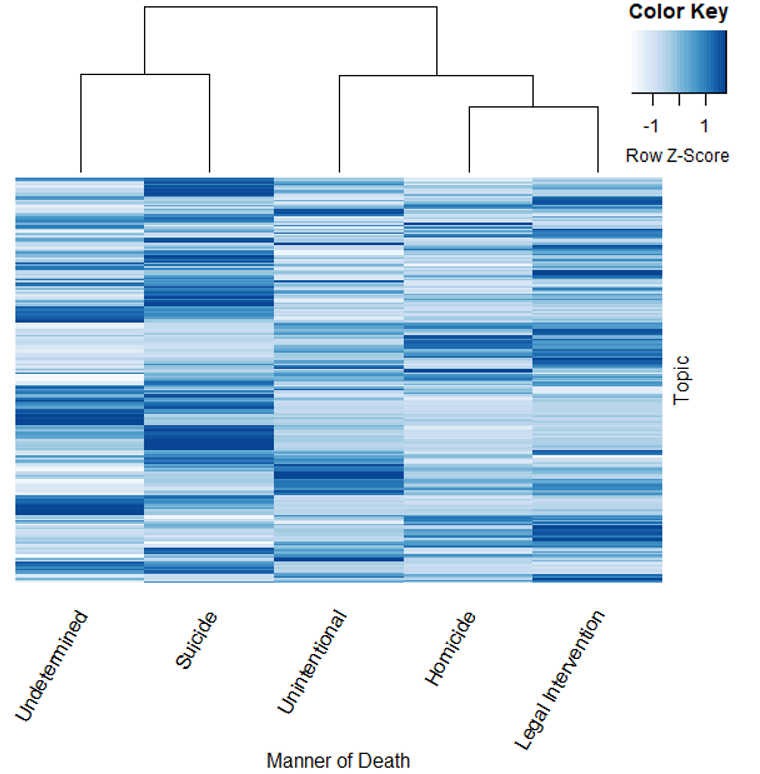}
\caption*{\textit{Note:} Each row represents the fraction of narratives with a given topic by manner of death, row standardized across all manners of death.}
\end{figure}

Figure \ref{figure:Heatmap of Topics by MOD} illustrates the prevalence of our 225 topics as patterned by manner of death (suicide, unintentional firearm, undetermined, homicide, and homicide by legal intervention). The dendogram shows that across manners of death, suicides are most similar in topic distributions to undetermined deaths; this makes sense, because many deaths may look like suicide but lack the required evidence for classification as suicide \citep{rosenberg1988operational}. It also shows that homicides are most similar to legal intervention deaths; this reflects the fact that legal intervention deaths are a unique \textit{type} of homicide \cite{barber2016homicides}.

\subsection*{Topics and Latent Gendered Dimensions}

Because the corresponding atom vectors live in an embedding space, we can apply common word embedding methods to our topics.  One prominent deductive approach uses knowledge about cultural connotations to extract a corresponding dimension in the semantic space. We extract a dimension for gender (masculine vs feminine) in the corpus, following standard word embedding methods \citep[e.g.,][]{bolukbasi2016quantifying}.\footnote{To extract a gender dimension, we average the vectors for the words: woman, women, female, females, she, her, herself, and hers. We then subtract out the average of the vectors for the words: man, men, male, males, he, him, himself, and his \citep{arseniev2020machine}.}  We then examine the topics that load most highly onto the gender dimension (i.e., have the highest or lowest cosine similarity). Cosine similarity can range from -1 to 1: for gender, the topics with large negative cosine similarity are more distinct to language about men (and not women), while topics with large positive cosine similarities are more distinct to language about women (and not men).

In our data, the most masculine topic is about the military, followed by topics about rural outdoor areas, rifles and shotguns, specific outdoor locations, and characteristics of suspects. The most feminine topic, by contrast, is one about sedative and pain medications, followed by topics about poisoning, children, drug concentrations, and psychiatric medications. 

\begin{figure}[tbhp]
\caption{Latent Gendered Meanings of Topics vs Prevalence of Topics in Female vs Male Decedents' Narratives.}\label{figure:Gendered Topics Valid}
\includegraphics[scale=0.65]{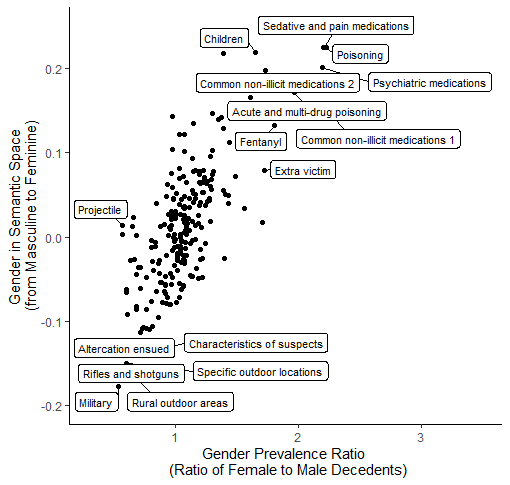}
\caption*{\textit{Notes:} N= 225 topics. For clarity, labels are shown only for topics with high or low gender meanings or gender prevalence ratios. The y-axis represents cosine similarity between a given topic and the gender dimension in semantic space. The x-axis represents the ratio of female decedents' narratives containing a given topic compared to narratives of male decedents. Spearman $\rho$ = 0.69 (p<0.0001).}
\end{figure}

We further compare the cosine similarity of topics to this gender dimension with the mean prevalence of each topic among female victims (versus among male victims) (figure \ref{figure:Gendered Topics Valid}). The strong correlation (Spearman $\rho$ = 0.69, p<0.0001) suggests that topics are gendered in semantic space in a way that corresponds to the gender of the victims in the narratives.

Next, we describe two topics in depth. Each has a high cosine similarity to the gender dimension. We select these topics because they are the most masculine and feminine topics (respectively) that describe weapons of death; weapon use has a well-known gendered pattern in violent death \citep[e.g.,][]{fox2017gender, callanan2012gender}. For each topic, we describe the most representative words and the case that loads most highly onto it. Then we use logistic regression to describe correlates of the topic: decedent demographics, manner of death, and number of decedents in the incident, controlling for word count.

\subsubsection*{Topic 141: Rifles and Shotguns}
Topic 141 reflects characteristics of long guns (e.g., rifles and shotguns). These firearms are typically owned for hunting and sport shooting \citep{hepburn2007us} and may be used to shoot at far ranges (compared to handguns). The most representative terms for this topic (by cosine similarity) are ``savage\_arms'', ``bolt\_action'', ``savage'', ``pump\_action,'' ``mossberg'', ``stoeger'', and ``stevens''. These terms refer to makes and models of long guns, as well as characteristics of gun action: how the gun is loaded and fired. The highest loading case describes the death of a young man accidentally shot by a friend playing with a rifle, who believed it was unloaded. The gun is described in depth (e.g., as a ``bolt action deer rifle''). The immediate cause of death was ruled as homicide. 

Topic 141 is the third most masculine topic in semantic space. This strong gender connotation reflects the fact that violent death by firearm typically involves males \citep{fox2017gender, callanan2012gender}. Logistic regression confirms that Topic 141 is distinctly more common among male victims (than females), controlling for characteristics listed in table \ref{table:regressiontopics} (adjusted odds ratio 0.49, 95 \% CI: 0.48-0.51). The strong gendered associations of this \textit{particular} gun-related topic in semantic space (compared to say, Topic 61: Handguns) could be because far more men than women own long guns \citep{hepburn2007us}. 

We also observe patterns in Topic 141 across other covariates. While prior work suggests that the majority of firearm-related decedents are Black \citep{goldstick2021current}, our results (table \ref{table:regressiontopics}) suggest that patterns may be more nuanced for deaths involving long guns. For instance, this topic is more common among American Indian/Alaska Native decedents, and less common in all other race/ethnicity groups (compared to Whites). And finally, the topic was more commonly present in incidents where there were multiple deaths, as one would see in mass shootings. Thus, more generally, this topic underscores the need for work on specific guns \citep[e.g.,][]{dresang2001gun, hanlon2019type, nestadt2017urban} in order to more effectively target prevention efforts aimed at firearm control.

\subsubsection*{Topic 53: Sedative and Pain Medications}
Topic 53 involves sedatives and medications that can be used to control pain. The most representative terms for this topic (by cosine similarity) include: ``phenergan'', ``motrin'', ``ultram,'' ``flexeril,'' ``endocet'', ``lunesta'', and ``hydrocodone\_apap.'' The highest loading case describes a middle-aged white male decedent who was found dead next to various prescription bottles with pain medications (e.g., methadone and hydrocodone). The immediate cause of death was ruled as suicide. In general, we found many topics focused on distinct groups of medications and drugs, attesting to the depth and patterned ways in which substances are described in the narratives. 

Topic 53 is also the most feminine topic in semantic space. This strong feminine connotation may reflect the fact that women are more likely to die by poisoning in suicides, the most common type of violent death for women \citep{ertl2019surveillance}. Logistic regression confirms that Topic 53 is distinctly more common among female victims, controlling for characteristics listed in table \ref{table:regressiontopics} (adjusted odds ratio= 2.52, 95\% CI: 2.47-2.58). We observe additional patterns of topic prevalence across these correlates. Compared to suicides, this topic is more common in undetermined deaths, but less common in all other deaths. The fact that unclassified deaths disproportionately involve this topic in their narratives resonates with broader scholarship on understanding the misclassification of death. Undetermined deaths are predominately associated with drug intoxication and poisoning \citep{rockett2018discerning, breiding2006variability} and many of the undetermined deaths involving drugs may be uncounted suicides \citep[e.g.,][]{donaldson2006classifying, stone2017deciphering}.

\setlength{\tabcolsep}{1pt}
\renewcommand{\arraystretch}{0.75}

\begin{table}[tbhp]
\caption{Characteristics of Violent Deaths with Two Selected Topics}\label{table:regressiontopics}
\begin{tabular}{@{}lcc@{}}
\toprule
  & \multicolumn{2}{@{}c}{\textbf{\textit{Topic}}} \\\midrule
  
    &  \textbf{Rifles and} & \textbf{Sedative and Pain} \\ 
    & \textbf{Shotguns} & \textbf{Medications}  \\ 
    Characteristic &  AOR (95\% CI)  & AOR (95\% CI) \\\midrule
    \textbf{Female Decedent$^1$} &  0.49  (0.48-0.51) & 2.52  (2.47-2.58)         \\
    \multicolumn{3}{@{}l}{\textbf{Decedent Race/Ethnicity$^2$}}\\
    \hspace{2mm} American Indian/Alaska Native, NH   &  1.31  (1.20-1.42) &  0.46  (0.41- 0.52)      \\
    \hspace{2mm} Asian/Pacific Islander, NH &   0.48  (0.43-0.54)  &  0.64  (0.59-0.70)      \\
    \hspace{2mm} Black or African American, NH  &   0.88  (0.85-0.91)  &  0.54  (0.51- 0.56)      \\
    \hspace{2mm} Hispanic    & 0.59  (0.56-0.62)  &   0.63  (0.60-0.67)    \\
    \hspace{2mm} Two or more races, NH    & 1.01  (0.92-1.10) &   0.80  (0.73-0.88)     \\
    \hspace{2mm} Unknown race, NH &   0.70  (0.56-0.87)   &  0.70  (0.56-0.87)    \\
  
   \multicolumn{3}{@{}l}{\textbf{Decedent Age (Years) $^3$}}\\
    \hspace{2mm} 20-29   &  0.96  (0.91-1.00) &   1.37  (1.29-1.46)  \\
    \hspace{2mm}  30-39    &  0.90  (0.86-0.95) &   1.74  (1.64-1.85)  \\
    \hspace{2mm}  40-49    &   0.93  (0.88-0.98)  &   1.97  (1.86-2.10)    \\
    \hspace{2mm}  50-59    &  1.03  (0.98-1.08)  & 2.17  (2.04-2.30)     \\
    \hspace{2mm}  60+     &   1.40  (1.33-1.47)  & 1.68  (1.58-1.79)        \\
    \multicolumn{3}{@{}l}{\textbf{Manner of Death$^4$}}\\
    \hspace{2mm} Homicide   &  0.79  (0.77-0.82) &   0.14  (0.13-0.15)    \\
    \hspace{2mm} Legal Intervention   &  1.09  (1.01-1.17) &    0.22  (0.19-0.26)     \\
    \hspace{2mm} Undetermined   &  0.06  (0.06-0.07)  &  2.01  (1.95-2.07)     \\
    \hspace{2mm} Unintentional   &   3.16  (2.84-3.51)  &  0.13  (0.10-0.19)    \\
    \textbf{Multiple Decedents in Incident$^5$}   & 1.76  (1.68-1.84) &  0.40  (0.37-0.43)\\
    \textbf{Word Count$^6$}  &    1.00  (1.00-1.00) & 1.00  (1.00-1.00) \\
  \bottomrule   
  \multicolumn{3}{p{8.5cm}}{\textbf{Notes:} N = 272,964 decedents. Topics are coded as present in any amount or not (1/0) in either the narrative of law enforcement reports or of medical examiner/coroner reports. AOR = Adjusted Odds Ratio; CI = Confidence Interval, NH=Non-Hispanic. 
$^1$Referent = Male.
$^2$Referent = non-Hispanic White.
$^3$Referent = 12-19.
$^4$Referent = Suicide.
$^5$Referent= Incidents with a single decedent.
$^6$This is the combined word count of both narratives.} \\
\end{tabular}
\end{table}

\section*{Discussion}

In this paper, we introduced a new method to model topics: Discourse Atom Topic Modeling (DATM). In DATM, topics, sentences, words, and other semantic structures are all represented in a single semantic space. Raw text can be mapped into this space to code individual documents as sequences of topics and thus draw out the prevalence of topics in a corpus. Using DATM, we discovered a range of themes buried in descriptions of lethal violence from a large administrative health dataset. We observed that the gendering of these topics in semantic space corresponds to the ratio of female versus male victims whose narratives mention these topics. We analyzed two highly gendered topics in depth (long guns and sedative and pain medications). Methodologically, our model builds on theoretical work to explain word embeddings and represent sentences in embedding spaces \citep{arora2016simple,arora2016latent,arora2018linear}, as well as a wealth of previous models to extract topics \citep[e.g.,][]{griffiths2004finding, blei2012probabilistic}. 

For computational social science and natural language processing, our DATM method opens new doors to discover patterns in large-scale text data. As a topic model, DATM picks up fine-grained, interpretable topical structures, as we have illustrated. Note that these topics are coherent despite the fact that no stopwords were pre-specified. This makes DATM ideal for real-world applications of text analysis, which are often domain specific. Further, DATM offers a cohesive, theoretically-motivated approach to \textit{integrate} questions that are often asked of topic models with questions often asked of embedding methods. A researcher can now ask, for example, not only what topics are in a corpus, but how these topics lie on latent semantic dimensions such as gender or social class. 

For public health, our results illustrate a range of patterns encoded in large-scale narrative data about suicides, homicides, and other violent deaths. Using DATM on these data offers a new way to break out of the well-worn categorical systems by which we interpret and monitor lethal violence.
We found that unstructured text data can hide potential patterns or trends that are not yet part of our standardized menu of structured variables. Such patterns could suggest new lines of research that aim to reduce violent death; for example, discovering new indicators of suicide risk, with eventual implications for medical providers or hotlines. Despite the wide use of the NVDRS for research and policy about lethal violence, actionable information in its text data remains largely out of reach. We hope that Discourse Atom Topic Modeling will provide an interpretable, flexible, theoretically grounded, and effective tool for scientists to unlock the potential of important datasets like the NVDRS.

\matmethods{Our data are drawn from the NVDRS from 2003 to 2017 \citep{paulozzi2004cdc}.  The NVDRS database included information about violent deaths forwarded from 34 U.S. states and the District of Columbia, for decedents aged 12 and older. This state-level information is abstracted into the NVDRS by public health workers (PHW) using a fully standardized codebook. These data include 307,249 violent deaths such as suicides, homicides, deaths of undetermined intent, unintentional deaths (primarily, shootings), and deaths related to legal intervention. We use two free-form text variables in the NVDRS written by PHW: narratives of 1) law enforcement reports and 2) medical examiner or coroner investigative reports. Death records may include one of these variables, both, or none, for a total of 568,262 narratives. As part of preprocessing, we transform commonly occurring phrases into single words (i.e., terms) \citep{rehurek_lrec}. We train our word embedding on all of these narratives using word2vec \citep{mikolov2013efficient, rehurek_lrec}. We limit our empirical investigation of topic distributions to the 272,979 (88.85\%) deaths which have at least 50 terms in either of the narratives. We further exclude 18 deaths where manner of death is missing or is coded as terrorism, leaving our final sample of 272,964 deaths. For details, see the Supplementary Information.

}

\showmatmethods{} 

\section*{Bibliography}
\bibliography{pnas-sample}

\end{document}